\documentclass[sigconf]{acmart}

\AtBeginDocument{%
  }

\copyrightyear{2026}
\acmYear{2026}
\setcopyright{cc}
\setcctype{by}
\acmConference[SA Technical Communications '26]{SIGGRAPH Asia 2026 Technical Communications}{December 01--04, 2026}{Kuala Lumpur, Malaysia}
\acmBooktitle{SIGGRAPH Asia 2026 Technical Communications (SA Technical Communications '26), December 01--04, 2026, Kuala Lumpur, Malaysia}
\acmDOI{10.1145/3829339.3847838}
\acmISBN{979-8-4007-2841-9/2026/12}

\usepackage{wrapfig}
\usepackage{acronym}
\usepackage{multirow}
\usepackage{soul}
\usepackage{xcolor}
\sethlcolor{green!30}
\usepackage{bibunits}
\defaultbibliographystyle{ACM-Reference-Format}

\acrodef{FFT}{Fast Fourier Transform}
\acrodef{rPPG}{Remote Photoplethysmography}
\acrodef{PPG}{Photoplethysmography}
\acrodef{2D}{two-dimensional}
\acrodef{3D}{three-dimensional}
\acrodef{GS}{Gaussian Splatting}
\acrodef{SH}{spherical harmonics}
\acrodef{MLP}{multilayer perceptron}

\newcommand{\heartian}{\includegraphics[height=0.8em]{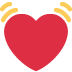} }

\begin{document}
\begin{bibunit}

\title[Heartian: Physiology-Aware Relightable Gaussian Head Avatar]{\protect\includegraphics[height=1em]{1f493.png} Heartian: Physiology-Aware Relightable Gaussian Head Avatar}
\titlenote{\protect\includegraphics[height=1em]{1f493.png}: \href{https://github.com/twitter/twemoji}{Twemoji} \copyright\ Twitter, Inc.\ and other contributors, 2019.}


\author{Xiaoyue Fan}
\affiliation{%
  \institution{University College London}
  \city{London}
  \country{UK}
}
\email{merryxyfan@gmail.com}

\author{Jose Echevarria}
\affiliation{%
  \institution{Adobe Research}
  \city{New York}
  \state{NY}
  \country{USA}}
\email{echevarr@adobe.com}

\author{Akshay Paruchuri}
\affiliation{%
  \institution{Stanford University}
  \city{Stanford}
  \state{CA}
  \country{USA}}
\email{akshaypa@stanford.edu}

\author{Kaan Akşit}
\affiliation{%
  \institution{University College London}
  \city{London}
  \country{UK}}
\email{kaanaksit@kaanaksit.com}

\settopmatter{authorsperrow=4}

\renewcommand{\shortauthors}{Fan et al.}
\newcommand{\skingaussian}{\mathcal{G}^s}
\newcommand{\fullgaussian}{\mathcal{G}}

\begin{abstract}
Gaussian head avatars typically model intrinsic facial appearance as temporally static, omitting subtle cardiac-induced skin-color variation. We propose \heartian Heartian, a physiology-aware modulation framework that learns cardiac-cycle-dependent per-frame albedo modulation of facial skin-region Gaussians within a relightable head avatar to encode remote photoplethysmography (rPPG) signals. Using synchronized contact PPG supervision, \heartian Heartian models the prescribed cardiac waveform as the sum of two Gaussian functions and learns per-frame spatial residuals via a lightweight MLP. Across 152 stationary recordings from UBFC-rPPG, PURE, and MMPD, attribute-space recovery of the supplied signal achieves a pooled recording-level heart-rate MAE of 0.29 bpm and MAPE of 0.38\%. The signals remain detectable after rendering by benchmark rPPG methods, with the best tested configuration -- a motion-augmented TS-CAN decoder pretrained on UBFC-rPPG -- recovering heart rate from the rendered MMPD avatars at 0.97 bpm MAE and 1.21\% MAPE. Meanwhile, \heartian Heartian maintains reconstruction quality comparable to the baseline, with negligible average PSNR degradation of 0.005 dB. Overall, our work embeds recoverable rPPG signals as controllable material attributes to subject-specific Gaussian head avatars, while retaining reconstruction quality.
\end{abstract}



\maketitle

\section{Introduction}

Conventional videos of people reveal physiological signals through subtle, temporal skin-color variation, from which rPPG recovers vital signs such as heart rate. Prior works synthesize physiological facial appearance in images~\cite{warmbodies} or videos~\cite{facevideo}, whereas ~\cite{lu2025rhythmgaussian} uses Gaussian representations to estimate unknown physiology from input video. However, 3D-consistent physiological appearance modeling remains underexplored in subject-specific, relightable head avatars.

Volumetric 3D human reconstruction has been significantly advanced by 3DGS ~\cite{kerbl3Dgaussians}. The work by ~\cite{lu2025rhythmgaussian} has validated the utility of GS in rPPG tasks with dynamic motions, where explicit 3D representation disentangles the chromatic and geometric variations beyond purely surface-pixel-level analysis. Meanwhile, HRAvatar~\cite{zhang2024hravatar} introduces a human head reconstruction method, leveraging Gaussian primitives upon an enhanced FLAME-based face model, supporting real-time relighting. However, it maintains a pose-independent static appearance, where per-Gaussian color properties are optimized globally rather than per frame. Consequently, while frame-wise lighting variations contribute to the final rendered appearance, inter-frame skin chromatic variations of physiological origin are inherently not captured. 

We propose \heartian Heartian, a physiology-aware modulation framework that embeds prescribed rPPG signals into relightable Gaussian head avatars. Building on HRAvatar reconstruction, we optimize per-frame spatial modulation factors acting on the albedo of facial skin-region Gaussians. Specifically, the modulation comprises two components: a fundamental waveform modeled as the sum of two Gaussian functions, following ~\cite{tang2020synthetic}, and a lightweight \ac{MLP} that learns per-frame spatial residuals. The rPPG signal is extracted as the mean green channel intensity after modulation across the skin-region Gaussians, supervised against the ground truth PPG waveform. In summary, our method enables the recovery of heart rate from embedded signals with a mean MAE of 0.29 bpm and MAPE of 0.38 \% against ground truth PPG measurements while preserving comparable reconstruction quality after modulation, with a marginal cost of 0.005 dB, 0.00003, and 0.0003 in average PSNR, SSIM, and LPIPS, respectively. Meanwhile, embedded rPPG signals are vastly preserved in the rendered videos, evaluated with the benchmark methods by rPPG-Toolbox~\cite{liu2022rppg}, exhibiting particularly strong performance on the MMPD dataset~\cite{MMPD2023}. Unlike methods that use a generalizable 4D Gaussian representation as an intermediate to disentangle feature variation for rPPG estimation~\cite{lu2025rhythmgaussian}, \heartian Heartian embeds prescribed rPPG signals via attribute modulation into subject-specific Gaussian head avatars as a renderable property. Furthermore, while retaining the relighting capability of HRAvatar, our pipeline enables rPPG avatar synthesis under novel illumination environments without compromising recoverable signal quality. Our strategy also enables prescribed waveform and heart-rate control within physiology-aware Gaussian representations, potentially supporting controlled physiological training augmentation and informing future physiology-aware real-time avatar models for applications such as telemedicine. \footnote{Our approach could be layered atop existing avatar infrastructure, such as teleconferencing pipelines (e.g., \href{https://dl.acm.org/doi/abs/10.1145/3641517.3664381}{\textit{Google Beam}}) or anatomically complete parametric head models (e.g., \href{https://arxiv.org/pdf/2607.23687}{\textit{GNM}}) to embed prescribed physiological signals into avatar reconstruction. Institutions such as \href{https://www.uni-hamburg.de/en/newsroom/presse/2022/pm20.html}{\textit{Universit\"at Hamburg}} have explored integrating vital-sign monitoring into telemedicine workflows, further suggesting that physiology-aware representations may eventually be incorporated into broader real-time avatar pipelines.}

\section{Methods}
Our method embeds synchronized rPPG signals into a relightable Gaussian head avatar reconstructed from monocular video, building on HRAvatar~\cite{zhang2024hravatar} as our baseline. \heartian Heartian is an offline, per-recording authoring method rather than a real-time physiology-inference system. Monocular videos undergo standard preprocessing and facial tracking to extract per-frame FLAME parameters, which are subsequently fed into HRAvatar to reconstruct a high-fidelity, subject-specific, and relightable head avatar represented as a set of physically parameterized 3D Gaussians $\fullgaussian$.

\paragraph{PPG Signals Modeling}

\begin{figure}[h]
    \centering
    \includegraphics[width=0.9\linewidth]{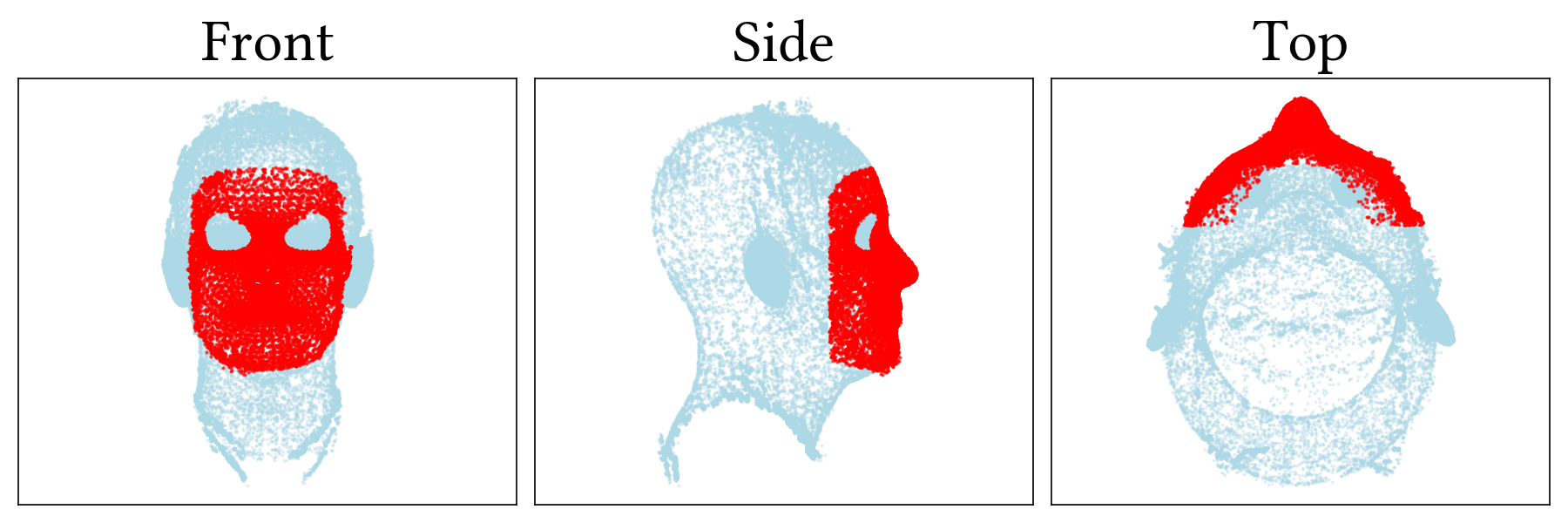}
    \caption{Skin Gaussians $\skingaussian$ (red) overlaid on $\fullgaussian$ (blue).}
    \label{fig:skin_gaussian}
\end{figure}

Cardiac pulsation results in skin color variations, imperceptible to the naked eye but detectable to cameras. We identify a dedicated set of Gaussians $\skingaussian$ from $\fullgaussian$ and modulate them during rendering, embedding the target rPPG signal without altering the underlying Gaussian parameters. To identify $\skingaussian$, we extract a facial skin-region mask from the well-trained $\fullgaussian$ representation. Each $\skingaussian_i$ is examined by its Euclidean distance to the nearest vertex on the FLAME mesh. As depicted in Figure~\ref{fig:skin_gaussian}, a selected $\fullgaussian_i$ is designated as $\skingaussian_i$ if this distance falls below a threshold (e.g., $0.02$) and its normal points in the front-facing direction, excluding Gaussians on the sides of the head where hair occlusion is likely.

\setlength{\intextsep}{6pt}

\begin{figure}[h]
    \centering
    \includegraphics[width=0.8\linewidth]{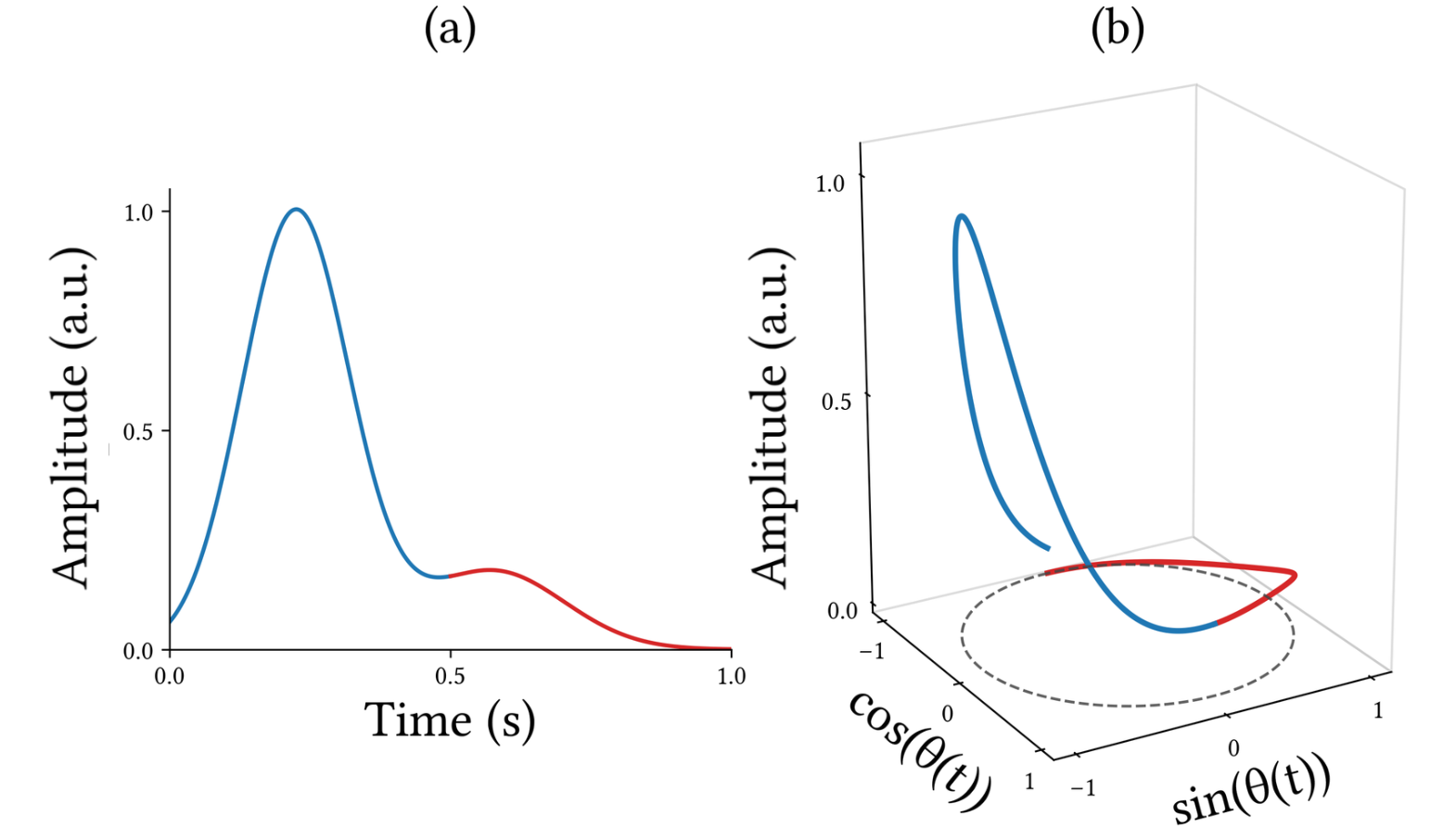}
    \caption{A typical PPG waveform with systolic (blue) and diastolic waves (red) in the time (a) and phase domains (b).}
    \label{fig:ppg_demo}
\end{figure}

\noindent For each $\skingaussian_{i}$, a spatially varied transient offset is applied to the green channel albedo $\mathbf{c}$ on a per-frame basis, while all other attributes remain unchanged. Specifically, the modulation at frame $t$ follows:
\begin{equation}
\mathbf{c}_i^{\text{m}}(t) = \mathbf{c}_i^{\text{base}}(t) + A \cdot m_i(t),
\end{equation}

\noindent where $A$ is a learnable scaling factor and $m_i(t)$ denotes the per-frame modulation scalar. The PPG waveform within each cardiac cycle characteristically exhibits two distinct peaks, a systolic and a diastolic wave. Their structure can be embedded in phase space by mapping the temporal signal to a unit circle parameterized by the cumulative cardiac phase $\theta(t)$~\cite{tang2020synthetic}, as illustrated in Figure~\ref{fig:ppg_demo}. Therefore, we model the $m_i(t)$ comprising the physiologically motivated fundamental waveform and the learned residual:
\begin{equation}
m_i(t) = \underbrace{\sum_{k=1}^{2} A_k \exp\left(-\frac{d(\theta(t), \mu_k)^2}{2\sigma_k^2}\right)}_{\text{fundamental}} +\ B \cdot \underbrace{f_{\text{MLP}}(\theta(t), b(t), p_i)}_{\text{residual}},
\end{equation}

\noindent where $d(\theta, \mu) = \arctan2(\sin(\theta - \mu), \cos(\theta - \mu))$ denotes the shortest angular distance on the unit circle. The fundamental component models the characteristic PPG waveform morphology as a sum of two Gaussian functions in phase space~\cite{tang2020synthetic}, with learnable centers $\mu_k$, widths $\sigma_k$, and amplitudes $A_k$. $\theta(t)$ is modeled as a learnable initial phase $\phi_0$ accumulated with per-frame increments $\{\delta_k\}$, which are passed through a softplus activation to ensure positivity, and projected onto the unit circle as:
\begin{equation}
\begin{aligned}
    \phi(t) &= \phi_0 + \sum_{k=0}^{t} \delta_k,\\
    \theta(t) &= \arctan2(\sin(\phi(t)), \cos(\phi(t))).
\end{aligned}
\end{equation}
The residual term is a lightweight $f_{\text{MLP}}$ conditioned on $\theta(t)$, a beat index embedding $b(t)$, and the normalized spatial position $p_i$ of each $\skingaussian_{i}$. While $\theta(t)$ encodes the continuous phase position within a cardiac circle, $b(t)$ provides a discrete beat identity that enables $f_{\text{MLP}}$ to capture inter-beat variability. The spatial conditioning on $p_i$ further allows the modulation to vary across the face, reflecting the spatially heterogeneous nature of the skin perfusion. Specifically, the phase embeddings follow the Fourier-based positional embedding approach~\cite{NeRF2020} with 10 harmonics:
\begin{equation}
e_{\phi}(\theta) = [\sin(k\theta), \cos(k\theta)]_{k=1}^{10} \in \mathbb{R}^{20}.
\end{equation}
\noindent $b(t)$ is encoded via a learnable embedding $e_b(t) \in \mathbb{R}^{8}$, while $p_i$ is embedded by a two-layer encoder into $e_s(p_i) \in \mathbb{R}^{8}$. The global features $[e_{\phi}(\theta),\, e_b(t)]$ are expanded and concatenated with the per-Gaussian $e_s(p_i)$, forming $\boldsymbol{x} = [e_{\phi}(\theta),\, e_b(t),\, e_s(p_i)]\in \mathbb{R}^{36}$, projected to dimension 128, passed through 2 residual blocks with tanh activations, and linearly projected to a per-Gaussian scalar residual.

\paragraph{Heart Rate Estimation Protocol}
Following the rPPG principle, heart rate is estimated from the temporal variation of the mean green albedo $\bar{c}$ across $\skingaussian$. Unlike RGB-rendered values, which are susceptible to environmental illumination variations and rendering noise, albedo provides a more stable and cleaner basis for extracting subtle signals. Functions of rPPG-Toolbox~\cite{liu2022rppg} are adapted for signal detrending, bandpass filtering, and frequency analysis via \ac{FFT}. The raw signal is detrended to remove low-frequency drift via a smoothness prior, followed by a first-order Butterworth bandpass filter to isolate cardiac frequencies ($f_L = 0.75$~Hz, $f_H = 3.0$~Hz) using zero-phase forward-backward filtering to prevent phase distortion. Heart rate is taken as the dominant spectral peak within the filter range via \ac{FFT}.



\paragraph{Optimization}
Our training follows a two-stage strategy. In the first stage, the HRAvatar baseline~\cite{zhang2024hravatar} is trained to optimize the full Gaussian set $\fullgaussian$ and FLAME parameters under $\mathcal{L}_{\text{base}}$, comprising the photometric loss. In the second stage, all baseline parameters are frozen to preserve well-learned geometry and static appearance, while only the per-frame-varying \ac{rPPG} signal modulation parameters of $\skingaussian$ are optimized. The rendered images pass through the full $\mathcal{L}_{\text{base}}$ to ensure the modulated albedo remains visually consistent. In addition, we incorporate rPPG-specific losses, targeting both frequency accuracy and waveform fidelity. The phase loss regularizes the cardiac rhythm by measuring the mean circular distance between $\theta(t)$ and the beat-aligned reference phase $\theta_{\text{gt}}$, constructed by detecting systolic peaks in the ground truth PPG signal and linearly interpolating between consecutive peaks. This normalization ensures each cardiac cycle occupies the same phase interval, allowing the global $\mu_k$ and $\sigma_k$ to learn a consistent morphological representation across the full sequence without drifting with heart rate variability:
\begin{equation}
\mathcal{L}_{\text{phase}} = \mathbb{E}\left[\left(\text{atan2}\left(\sin(\theta - \theta_{\text{gt}}),\ \cos(\theta - \theta_{\text{gt}})\right)\right)^2\right].
\end{equation}
Besides, we apply a weak phase regularization via instantaneous phase estimates 
$\theta_{\text{POS}}$ derived from raw frames using the Plane-Orthogonal-to-Skin (POS) algorithm~\cite{POS2016} followed by a Hilbert transform. Unlike the supervised $\mathcal{L}_{\text{phase}}$ enforcing morphological consistency, POS-generated signals offer stable  periodicity as a soft prior to discourage arbitrary phase drift while preserving the model's capacity to learn morphological variation:
\begin{equation}
\mathcal{L}_{\text{pos}} = \mathbb{E}\left[\left(\text{atan2}\left(\sin(\theta - \theta_{\text{pos}}),\ \cos(\theta - \theta_{\text{pos}})\right)\right)^2\right].
\end{equation}
The waveform losses are applied along two separate supervision paths using the same ground truth PPG signal. The MSE loss $\mathcal{L}_{\text{mse}}$ and Pearson correlation loss $\mathcal{L}_{\text{pearson}}$ supervise the fundamental signal directly, encouraging the Gaussian parameters to approximate the target waveform morphology. The same losses are also applied to the full modulation signal, providing gradients exclusively to $f_{\text{MLP}}$ for correcting the remaining spatially varying discrepancy:
where $s$ denotes the detrended and normalized input signal for each respective path. To ensure physical plausibility, we additionally apply constraints on the fundamental waveform parameters following~\cite{tang2020synthetic}, enforcing $1\geq A_1>A_2\geq0$, $2\geq\sigma_2>\sigma_1\geq0$, and $\pi\geq\mu_2>\mu_1\geq-\pi$ via quadratic penalty terms:
\begin{equation}
\mathcal{L}_{\text{reg}} = \lambda_{\text{A}}\mathcal{L}_{A} + \lambda_{\sigma}\mathcal{L}_{\sigma} + \lambda_{\mu}\mathcal{L}_{\mu}.
\end{equation}
The total loss in the second stage is:
\begin{equation}
\mathcal{L} = \mathcal{L}_{\text{base}} 
+ \lambda_{\text{pha}}\mathcal{L}_{\text{phase}} 
+ \lambda_{\text{pos}}\mathcal{L}_{\text{pos}} 
+ \lambda_{\text{m}}\mathcal{L}_{\text{mse}} 
+ \lambda_{\text{p}}\mathcal{L}_{\text{pearson}} 
+ \mathcal{L}_{\text{reg}},
\end{equation}
where $ \lambda_{\text{pha}} = \lambda_{\text{m}} = \lambda_{\text{p}} = 1$, $\lambda_{\text{A}} = \lambda_{\sigma} = 0.6$, $\lambda_{\text{pos}} = 0.3$ and $\lambda_{\mu} = 0.1$.

\section{Implementation}

\paragraph{Our pipeline} We base our pipeline on HRAvatar~\cite{zhang2024hravatar} using PyTorch and optimize with Adam on an RTX $4060$ GPU. Training proceeds in two stages: the first stage follows the same configuration to generate the baseline avatar, while the second stage introduces and trains the rPPG modulation parameters for $60$ epochs. More details are provided in the supplementary material.

\paragraph{Dataset} We evaluate on three datasets: UBFC-rPPG ~\cite{bobbia2017unsupervised}, PURE~\cite{PURE2014}, and MMPD~\cite{MMPD2023}. For PURE and UBFC-rPPG, we select 10 subjects from each, all recorded under stationary motion. For MMPD, the full 33 subjects are utilized, spanning Fitzpatrick skin types 3 to 6, consistent with stationary motion and retaining all four lighting conditions, yielding 132 videos from this dataset and 152 videos in total. All monocular videos are extracted into frames, cropped to a face-centered 512 $\times$ 512 px region, and foreground-matted across the full sequences.

\section{Evaluation and Discussion}

\begin{table}[t]
\centering
\caption{Benchmark Evaluation Results. Performance of benchmark methods on our rendered \heartian Heartian videos, evaluated across datasets. Green highlights indicate metrics where embedded signals remain recoverable on par with the original benchmark results on the same selected subsets.}
\label{tab:benchmark}
\resizebox{\columnwidth}{!}{
\begin{tabular}{ll cc cc cc}
\toprule
& & \multicolumn{6}{c}{\textbf{Test Set}} \\
\cmidrule(lr){3-8}
& & \multicolumn{2}{c}{UBFC-rPPG} & \multicolumn{2}{c}{PURE} & \multicolumn{2}{c}{MMPD} \\
\cmidrule(lr){3-4} \cmidrule(lr){5-6} \cmidrule(lr){7-8}
\textbf{Method} & \textbf{Train Set} & MAE$\downarrow$ & MAPE$\downarrow$ & MAE$\downarrow$ & MAPE$\downarrow$ & MAE$\downarrow$ & MAPE$\downarrow$ \\
\midrule
\textsc{POS} & \multicolumn{1}{c}{-} & \hl{1.23 $\pm$ 0.77} & \hl{1.01 $\pm$ 0.62} & 14.23 $\pm$ 6.78 & 29.34 $\pm$ 13.98 & \hl{1.53 $\pm$ 0.63} & \hl{2.66 $\pm$ 1.20} \\
\midrule
\multirow{2}{*}{TS-CAN}
  & UBFC-rPPG & - & - & 10.01 $\pm$ 4.52 & 17.89 $\pm$ 8.68 & \hl{2.59 $\pm$ 0.52} & \hl{3.89 $\pm$ 0.94} \\
  & PURE      & 9.22 $\pm$ 7.47 & 7.41 $\pm$ 5.56 & - & - & \hl{2.51 $\pm$ 0.57} & \hl{3.87 $\pm$ 1.03} \\
\cmidrule(lr){2-8}
\multirow{1}{*}{TS-CAN (MA)}
  & UBFC-rPPG & - & - & \hl{4.74 $\pm$ 4.22} & \hl{5.23 $\pm$ 4.67} & \hl{0.97 $\pm$ 0.27} & \hl{1.21 $\pm$ 0.31} \\
\cmidrule(lr){2-8}
\multirow{2}{*}{PhysFormer}
  & UBFC-rPPG & - & - & 17.40 $\pm$ 6.55 & 34.29 $\pm$ 13.78 & \hl{8.69 $\pm$ 1.31} & \hl{13.96 $\pm$ 2.21} \\
  & PURE      & 5.53 $\pm$ 4.44 & 4.37 $\pm$ 3.32 & - & - & \hl{3.12 $\pm$ 1.08} & \hl{5.27 $\pm$ 1.95} \\
\cmidrule(lr){2-8}
\multirow{2}{*}{FactorizePhys}
  & UBFC-rPPG & - & - & 14.58 $\pm$ 6.82 & 29.95 $\pm$ 14.13 & \hl{2.88 $\pm$ 0.98} & \hl{4.83 $\pm$ 1.70} \\
  & PURE      & 6.41 $\pm$ 5.09 & 5.05 $\pm$ 3.80 & - & - & \hl{1.38 $\pm$ 0.50} & \hl{2.40 $\pm$ 0.99} \\
\midrule
\textbf{Ours vs Baseline} &  &  &  \\
\multirow{1}{*}{POS}
  & \multicolumn{1}{c}{-} & \textcolor{green!70!black}{- 21.97}  &  \textcolor{green!70!black}{- 18.88} & \textcolor{green!70!black}{- 10.72} &  \textcolor{green!70!black}{- 13.78} &  \textcolor{green!70!black}{- 12.72} & \textcolor{green!70!black}{- 17.78} \\
\cmidrule(lr){2-8}
\multirow{1}{*}{TS-CAN}
  & PURE & \textcolor{green!70!black}{- 26.98} & \textcolor{green!70!black}{- 24.72} & - & - & \textcolor{green!70!black}{- 12.87} &  \textcolor{green!70!black}{- 16.28}  \\
\cmidrule(lr){2-8}
\multirow{1}{*}{TS-CAN (MA)}
  & UBFC-rPPG & - & - & \textcolor{green!70!black}{- 15.73}  & \textcolor{green!70!black}{- 25.01}  & \textcolor{green!70!black}{- 13.31}  &   \textcolor{green!70!black}{- 17.28} \\
\bottomrule
\multicolumn{8}{c}{\footnotesize MAE = Mean Absolute Error in HR estimation (Beats/Min), MAPE = Mean Percentage Error (\%).}
\end{tabular}
}
\end{table}

\paragraph{Evaluation} We evaluate our method by reorganizing the rendered avatar videos according to the requirements of each benchmark, and subsequently testing them through rPPG-Toolbox ~\cite{liu2022rppg}. Our evaluation includes both the traditional unsupervised learning method POS~\cite{POS2016} and three cutting-edge pre-trained supervised neural networks - TS-CAN~\cite{TSCAN2020}, PhysFormer~\cite{Phys2022}, and FactorizePhys~\cite{Fac2024} - to assess the fidelity and recoverability of the prescribed rPPG signals embedded per recording. Since Gaussian-based avatar reconstructions inherently exhibit subtle frame-level jittering, we further evaluate \heartian Heartian with a motion-augmented variant of TS-CAN~\cite{MA2023} to investigate this impact. Table~\ref{tab:benchmark} shows \heartian Heartian preserves consistently recoverable rPPG signals after rendering across three datasets. Our method achieves particularly strong performance on MMPD~\cite{MMPD2023}, stably remaining recoverable by off-the-shelf rPPG estimators. In contrast, the best-performing method varies across the other two datasets, revealing gaps across datasets and benchmark methods. Notably, PURE~\cite{PURE2014} is more sensitive to mesh jittering, whose performance is substantially improved with motion-augmented TS-CAN~\cite{MA2023}, among others. Besides, PhysFormer~\cite{Phys2022} exhibits the largest deviation among supervised methods, as its transformer-based architecture generalizes less robustly across datasets with differing acquisition conditions~\cite{liu2022rppg}. We further composite the rendered avatar videos onto dynamic real backgrounds, showing no consistent degradation and confirming the robustness of our method beyond mere background removal. A cross-subject experiment on three subjects additionally demonstrates the controllability of our strategy, with the prescribed heart rates explicitly recovered. Beyond rPPG signals, we also assess the reconstruction quality of our modulated \heartian Heartian. A small number of videos exhibit visual artifacts due to initial transient spikes in the ground truth signal during data acquisition, which introduce large modulation magnitude, and lighting variation within a single recording, which heightens the inconsistent bases for albedo modulation strategy. However, these artifacts can be eliminated by adjusting the modulation scaling factor accordingly. Our method maintains reconstruction quality comparable to the baseline after albedo modulation with an average of 35.38 PSNR, 0.9697 SSIM, and 0.0374 LPIPS. Extensive results and visualizations are provided in the supplementary material.

\paragraph{Discussion}
Since rPPG estimation relies on subtle skin color variations, signals are sensitive to degradation during video compression or transmission. Our approach of embedding rPPG signals directly into the Gaussian albedo offers attribute-level signal preservation. Table~\ref{tab:attribute_signal} shows the metrics of rPPG signals extracted from our modulated \heartian Heartian compared with the static HRAvatar baseline, whose extracted albedo is consistently a flat line, demonstrating its lack of inter-frame variation and thus heart rate information. Compared with results derived from the rendered videos in Table~\ref{tab:benchmark}, the extracted signals show stronger fidelity, suggesting that rPPG information is numerically preserved within the avatar representation when it attenuates through rendering or post-processing.

\begin{table}[h]
\centering
\caption{Attribute-level Evaluation. rPPG signal metrics extracted from the Gaussian albedo of baseline and \heartian Heartian.}
\label{tab:attribute_signal}
\footnotesize
\begin{tabular*}{\linewidth}{@{\extracolsep{\fill}}l ccc ccc}
\toprule
& \multicolumn{3}{c}{\textbf{Baseline}} & \multicolumn{3}{c}{\textbf{Ours}} \\
\cmidrule(lr){2-4} \cmidrule(lr){5-7} 
\textbf{Dataset} & MAE & MAPE & SNR & MAE$\downarrow$ & MAPE$\downarrow$ & SNR$\uparrow$\\
\midrule
UBFC-rPPG & 59.00 & 53.81 & -21.52 & 0.00 & 0.00 & 1.45 \\
PURE & 21.53 & 25.37 & -7.13 & 0.17 & 0.27 & 3.47 \\
MMPD & 39.11 & 34.81 & -11.47 & 0.33 & 0.42 & 4.04 \\
\bottomrule
\end{tabular*}
\end{table} 

Lighting condition and skin tone have been challenging factors for rPPG signal estimation. However, benefiting from our attribute-level embedding strategy, well-learned signals remain largely embedded and recoverable across predefined external conditions in MMPD~\cite{MMPD2023}, as shown in Table~\ref{tab:mmpd}. Benchmark methods also demonstrate the same domain gaps in this comparison. The signals persist across skin tone 6 -- typically the most challenging skin tone -- as well as incandescent and natural lighting. Few biased cases predominantly occur for olive and mixed yellow-brown tones (i.e., skin tone 4 and 5). Since our direct modulation operates merely on the green channel, skin tones and lighting conditions that reduce green reflectance and thus the amplitude of the modulation base inherently weaken the modulation effects in our strategy. Hence, a similar effect is observed under warm and vibrant LED-high lighting. Furthermore, we conduct an ablation study on relighting effects by applying generated, evenly distributed white, cool, and warm lighting, as well as a real-world environment map captured in a hospital. Metrics remain largely consistent with those under original lighting with a mean $\Delta$MAE of $+0.20$ bpm, demonstrating the robustness of our method under the tested relit conditions. More details and future directions are in the supplementary material.

\begin{table}[t]
\centering
\caption{Lights and skin tones comparison within MMPD~\cite{MMPD2023}, tested with models pretrained on UBFC-rPPG~\cite{bobbia2017unsupervised}.}
\label{tab:mmpd}
\resizebox{\columnwidth}{!}{
\begin{tabular}{r cc cc cc}
\toprule
& \multicolumn{2}{c}{\textbf{POS}} & \multicolumn{2}{c}{\textbf{TS-CAN (MA)}} & \multicolumn{2}{c}{\textbf{FactorizePhys}}\\
\cmidrule(lr){2-3} \cmidrule(lr){4-5} \cmidrule(lr){6-7} 
& MAE $\downarrow$ & MAPE $\downarrow$ & MAE $\downarrow$ & MAPE$\downarrow$ & MAE $\downarrow$ & MAPE$\downarrow$ \\
\midrule
\textbf{Skin tone} &  &  &  &  \\
3 & 1.01 $\pm$ 0.82 & 1.80 $\pm$ 1.52 & 1.04 $\pm$ 0.44 & 1.35 $\pm$ 0.52 & 2.89 $\pm$ 1.48 & 4.66 $\pm$ 2.38 \\
4 & 1.91 $\pm$ 1.44 & 2.33 $\pm$ 1.82 & 2.17 $\pm$ 1.11 & 2.35 $\pm$ 1.11 & 2.89 $\pm$ 1.88 & 3.61 $\pm$ 2.41  \\
5 & 3.80 $\pm$ 2.34 & 7.44 $\pm$ 4.76 & 0.62 $\pm$ 0.34 & 0.87 $\pm$ 0.45 & 4.90 $\pm$ 2.89 & 9.50 $\pm$ 5.81\\
6 & 0.47 $\pm$ 0.21 & 0.53 $\pm$ 0.25 & 0.10 $\pm$ 0.10 & 0.12 $\pm$ 0.11 & 0.10 $\pm$ 0.07 & 0.12 $\pm$ 0.09 \\
\midrule
\textbf{Light} &  &  &  \\
LED-low  & 1.41 $\pm$ 0.94  & 2.42 $\pm$ 1.84 & 0.63 $\pm$ 0.24 &  0.81 $\pm$ 0.32 & 4.34 $\pm$ 2.69 & 6.91 $\pm$ 4.31\\
LED-high  & 3.40 $\pm$ 2.16 & 6.52 $\pm$ 4.24 & 0.37 $\pm$ 0.14 & 0.54 $\pm$ 0.21 & 3.30 $\pm$ 2.01 & 6.26 $\pm$ 3.97 \\
Incandescent& 0.42 $\pm$ 0.26 & 0.58 $\pm$ 0.38 & 0.98 $\pm$ 0.31 & 1.34 $\pm$ 0.41 & 2.26 $\pm$ 1.51 & 3.75 $\pm$ 2.66  \\
Nature& 0.90 $\pm$ 0.76 & 1.12 $\pm$ 0.96 & 1.89 $\pm$ 1.01 & 2.13 $\pm$ 1.12 & 1.65 $\pm$ 1.36 & 2.41 $\pm$ 2.06\\
\bottomrule
\end{tabular}
}
\end{table} 



\putbib[sample-base]
\end{bibunit}



\clearpage
\onecolumn
\pagestyle{empty}
\setcounter{section}{0}
\renewcommand{\thesection}{\arabic{section}}
\begin{bibunit}
\setcounter{table}{0}
\setcounter{figure}{0}
\section{Expanded Results}
\begin{figure}[h]
    \centering
    \includegraphics[width=\linewidth]{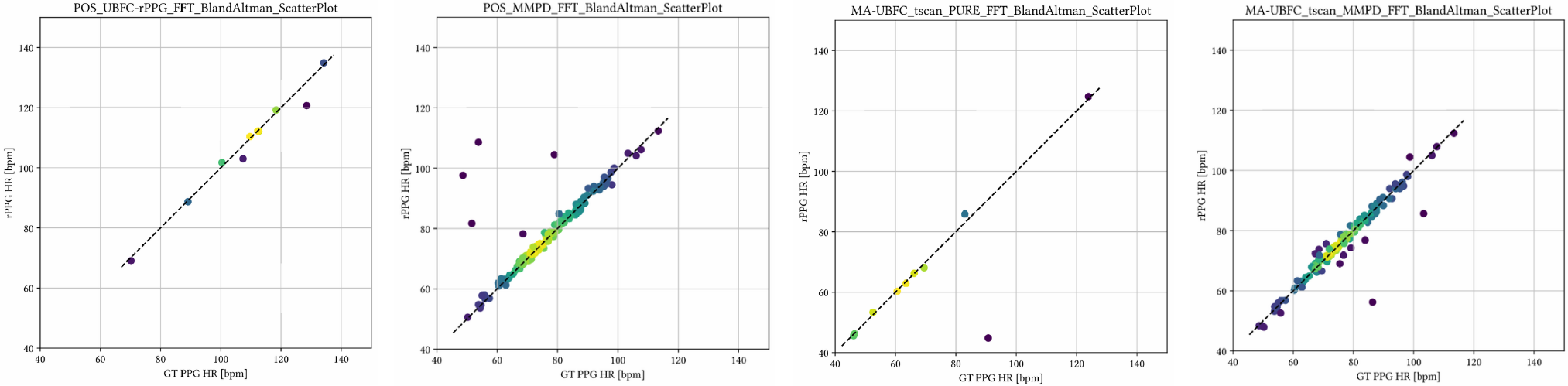}
    \vspace{-18pt}
    \caption{Visualization of Recovered Heart Rate Accuracy. Scatter plots of rPPG estimates from our rendered \heartian Heartian videos versus ground truth using POS for UBFC-rPPG and MMPD, and motion-augmented TS-CAN for PURE and MMPD.}
    \label{fig:BA}
\end{figure}

We present additional benchmark evaluation results using the rPPG-Toolbox~\cite{liu2022rppg}, utilizing unsupervised methods POS~\cite{POS2016} and pre-trained supervised neural networks - TS-CAN~\cite{TSCAN2020}, motion-augmented (MA) TS-CAN~\cite{MA2023}, PhysFormer~\cite{Phys2022}, and FactorizePhys~\cite{Fac2024} - shown in Table~\ref{tab:benchmark_supp} and representative scatter plots in Figure~\ref{fig:BA}.


\setlength{\intextsep}{6pt}
\begin{table}[h]
\centering
\caption{Additional Benchmark Evaluation Results. Performance of benchmark methods on our rendered \heartian Heartian avatar videos, evaluated across the UBFC-rPPG \cite{bobbia2017unsupervised}, PURE \cite{PURE2014}, and MMPD \cite{MMPD2023} datasets. Green highlights indicate metrics where signals remain recoverable on par with the benchmark results on the same subsets.}
\label{tab:benchmark_supp}
\vspace{-8pt}
\resizebox{\columnwidth}{!}{%
\begin{tabular}{ll ccccc ccccc ccccc}
\toprule
& & \multicolumn{15}{c}{\textbf{Test Set}} \\
\cmidrule(lr){3-17}
& & \multicolumn{5}{c}{UBFC-rPPG} & \multicolumn{5}{c}{PURE} & \multicolumn{5}{c}{MMPD} \\
\cmidrule(lr){3-7} \cmidrule(lr){8-12} \cmidrule(lr){13-17}
\textbf{Method} & \textbf{Train Set} 
  & MAE & RMSE & MAPE & $\rho$ & SNR
  & MAE & RMSE & MAPE & $\rho$ & SNR
  & MAE & RMSE & MAPE & $\rho$ & SNR \\
\midrule
POS & \multicolumn{1}{c}{-} & \hl{1.23 $\pm$ 0.77} & \hl{2.75 $\pm$ 2.43} & \hl{1.01 $\pm$ 0.62} & \hl{0.98 $\pm$ 0.05} & \hl{0.26 $\pm$ 0.92} & 14.23 $\pm$ 6.78 & 25.75 $\pm$ 18.04 & 29.34 $\pm$ 13.98 & 0.48 $\pm$ 0.30 & 2.93 $\pm$ 0.56 & \hl{1.53 $\pm$ 0.63} & \hl{7.43 $\pm$ 5.53} & \hl{2.66 $\pm$ 1.20} & \hl{0.82 $\pm$ 0.04} & \hl{4.20 $\pm$ 0.25} \\
\midrule
\multirow{2}{*}{TS-CAN}
  & UBFC-rPPG & - & - & - & - & - & 10.01 $\pm$ 4.52 & 17.46 $\pm$ 12.54 & 17.89 $\pm$ 8.68 & 0.68 $\pm$ 0.25 & -3.31 $\pm$ 0.66 & \hl{2.59 $\pm$ 0.52} & \hl{6.60 $\pm$ 4.60} & \hl{3.89 $\pm$ 0.94} & \hl{0.85 $\pm$ 0.04} & \hl{-1.71 $\pm$ 0.17}\\
  & PURE      & 9.22 $\pm$ 7.47 & 25.37 $\pm$ 24.62 & 7.41 $\pm$ 5.56 & 0.23 $\pm$ 0.34 & -4.9 $\pm$ 1.14 & - & - & - & - & - & \hl{2.51 $\pm$ 0.57} & \hl{7.02 $\pm$ 4.85} & \hl{3.87 $\pm$ 1.03} & \hl{0.83 $\pm$ 0.04} & \hl{-1.76 $\pm$ 0.16} \\
\cmidrule(lr){2-17}
\multirow{1}{*}{TS-CAN (MA)}
  & UBFC-rPPG & - & - & - & - & - & \hl{4.74 $\pm$ 4.22} & \hl{14.18 $\pm$ 13.80} & \hl{5.23 $\pm$ 4.67} & \hl{0.82 $\pm$ 0.19} & \hl{-0.28 $\pm$ 0.96} & \hl{0.97 $\pm$ 0.27} & \hl{3.35 $\pm$ 2.63}  & \hl{1.21 $\pm$ 0.31} & \hl{0.96 $\pm$ 0.02} & \hl{-0.40 $\pm$ 0.15} \\
\cmidrule(lr){2-17}
\multirow{2}{*}{PhysFormer}
  & UBFC-rPPG & - & - & - & - & - & 17.40 $\pm$ 6.55 & 27.06 $\pm$ 18.36 & 34.29 $\pm$ 13.78 & 0.50 $\pm$ 0.30 & -2.45 $\pm$ 0.69 & \hl{8.69 $\pm$ 1.31} & \hl{17.41 $\pm$ 7.87} & \hl{13.96 $\pm$ 2.21} & \hl{0.25 $\pm$ 0.08} & \hl{-2.62 $\pm$ 0.19}\\
  & PURE      & 5.53 $\pm$ 4.44 & 15.11 $\pm$ 14.60 & 4.37 $\pm$ 3.32 & 0.64 $\pm$ 0.26 & -2.34 $\pm$ 1.00 & - & - & - & - & - & \hl{3.12 $\pm$ 1.08} & \hl{12.80 $\pm$ 8.73} & \hl{5.27 $\pm$ 1.95} & \hl{0.58 $\pm$ 0.07} & \hl{0.18 $\pm$ 0.19} \\
\cmidrule(lr){2-17}
\multirow{2}{*}{FactorizePhys}
  & UBFC-rPPG & - & - & - & - & - & 14.58 $\pm$ 6.82 & 26.06 $\pm$ 18.21 & 29.95 $\pm$ 14.13 & 0.48 $\pm$ 0.30 & 0.58 $\pm$ 0.74 & \hl{2.88 $\pm$ 0.98} & \hl{11.69 $\pm$ 7.62} & \hl{4.83 $\pm$ 1.70} & \hl{0.62 $\pm$ 0.06} & \hl{1.70 $\pm$ 0.22} \\
  & PURE      & 6.41 $\pm$ 5.09 & 17.32 $\pm$ 16.77 & 5.05 $\pm$ 3.80 & 0.55 $\pm$ 0.29 & -1.26 $\pm$ 1.00 & - & - & - & - & - &\hl{1.38 $\pm$ 0.50} & \hl{5.91 $\pm$ 4.55} & \hl{2.40 $\pm$ 0.99} & \hl{0.88 $\pm$ 0.04} & \hl{1.66 $\pm$ 0.19} \\
\bottomrule
\multicolumn{17}{c}{\footnotesize MAE = Mean Absolute Error (BPM), RMSE = Root Mean Square Error (BPM), MAPE = Mean Percentage Error (\%), $\rho$ = Pearson Correlation, SNR = Signal-to-Noise Ratio (dB).}
\end{tabular}
}
\end{table}

\setlength{\intextsep}{0pt}


\setlength{\intextsep}{6pt}
\section{Implementation Details}
Facial skin Gaussians are selected based on their absolute distance to the nearest mesh vertex in canonical space, with a per-dataset threshold \{0.006, 0.01, 0.016\} for UBFC-rPPG~\cite{bobbia2017unsupervised}, PURE~\cite{PURE2014}, and MMPD~\cite{MMPD2023}, respectively, to account for scale and Gaussian density differences across optimized avatars by differing camera setups and capture conditions. For POS-based~\cite{POS2016} weak supervision, we adopt two configurations depending on dataset characteristics. For MMPD~\cite{MMPD2023}, which involves compressed mobile phone footage, we apply a 1.6s sliding window to obtain locally stable phase estimates. For the other two, which provide uncompressed or high-fidelity recordings, we adopt a simplified global POS formulation, as sliding windows tend to introduce spurious phase shifts. Regarding optimizer settings, all scheduled parameters follow an exponential decay: $\delta_k$ and $\phi_0$ decay over $60{,}000$ frame-level iterations from $9\times10^{-3}$ to $3\times10^{-4}$ and from $2\times10^{-3}$ to $5\times10^{-4}$, respectively, while $f_{\text{MLP}}$ decays over $80{,}000$ iterations from $3.8\times10^{-3}$ to $5\times10^{-4}$. The fundamental waveform parameters use fixed learning rates: $3\times10^{-4}$ for $A_1, A_2$; $8\times10^{-4}$ and $1\times10^{-3}$ for $\mu_1, \mu_2$; $6\times10^{-4}$ and $2\times10^{-3}$ for $\sigma_1, \sigma_2$; $1\times10^{-4}$ for $B$; and $9\times10^{-5}$ for $A$. We apply softplus activation to $A_1, A_2, \sigma_1,$ and $\sigma_2$ to enforce physical positivity and resist rapid shrinkage caused by photometric loss, which initially tends to suppress albedo modulation by driving them towards zero.

\setlength{\intextsep}{0pt}
\section{Ablation Study and Future Work}

\paragraph{Background effects}
The segmented foregrounds are composited onto a black background during avatar rendering by the default setting. To verify that our evaluation results are not confounded by the foreground matting, we add the temporal real backgrounds back onto the rendered avatars via alpha compositing as an ablation study. Under the same evaluation setting, metrics slightly change across methods and selected subsets, without a consistent degradation, as shown in Table~\ref{tab:bg}. In general, pretrained models were trained on data containing real backgrounds. Background composition naturally restores the global image statistics that pre-trained neural networks rely on after normalization. Specifically, PURE's~\cite{PURE2014} background is more complex and realistic, unlike the single-color backdrops of others. Consequently, models pretrained on PURE tend to benefit more from background composition. This effect also relates to the sensitivity to global spatial context across methods. Meanwhile, we noticed that motion-augmented TS-CAN~\cite{MA2023} consistently maintains strong and stable performance, highlighting its robustness to the temporal mesh jittering inherently introduced by avatar rendering. This points to a potential direction of combining motion augmentation with physiology-aware avatars to generate robust synthetic rPPG data, which could be incorporated into future systems to improve the real-world and camera-specific rendering for cross-dataset generalization.


\setlength{\intextsep}{6pt}
\begin{table}[h]
\centering
\caption{Background Compositing Ablation. Metric changes from background compositing over the default black background.}
\label{tab:bg}
\vspace{-10pt}
\resizebox{\columnwidth}{!}{%
\begin{tabular}{ll ccccc ccccc ccccc}
\toprule
& & \multicolumn{15}{c}{\textbf{Test Set}} \\
\cmidrule(lr){3-17}
& & \multicolumn{5}{c}{UBFC-rPPG} & \multicolumn{5}{c}{PURE} & \multicolumn{5}{c}{MMPD} \\
\cmidrule(lr){3-7} \cmidrule(lr){8-12} \cmidrule(lr){13-17}
\textbf{Method} & \textbf{Train Set} 
  & MAE & RMSE & MAPE & $\rho$ & SNR
  & MAE & RMSE & MAPE & $\rho$ & SNR
  & MAE & RMSE & MAPE & $\rho$ & SNR \\
\midrule
POS & \multicolumn{1}{c}{-} &  \textcolor{yellow!70!black}{+ 0.87} & \textcolor{red!70!black}{+ 1.15} & \textcolor{yellow!70!black}{+ 0.87} & \textcolor{yellow!70!black}{- 0.01} & \textcolor{yellow!70!black}{- 0.94} & \textcolor{yellow!70!black}{+ 0.52} & \textcolor{yellow!70!black}{+ 0.05} & \textcolor{yellow!70!black}{+ 0.62} & \textcolor{yellow!70!black}{- 0.01} & \textcolor{yellow!70!black}{- 0.22}  & \textcolor{red!70!black}{+ 3.04} & \textcolor{red!70!black}{+ 5.45} & \textcolor{red!70!black}{+ 5.62} & \textcolor{yellow!70!black}{- 0.24} & \textcolor{red!70!black}{- 2.09}\\
\midrule
\multirow{1}{*}{TS-CAN (MA)}
  & UBFC-rPPG & - & - & - & - & - & \textcolor{yellow!70!black}{0.00} & \textcolor{yellow!70!black}{0.00} & \textcolor{yellow!70!black}{0.00} & \textcolor{yellow!70!black}{0.00} & \textcolor{yellow!70!black}{+ 0.06} & \textcolor{yellow!70!black}{- 0.30} & \textcolor{green!70!black}{- 1.90}  & \textcolor{yellow!70!black}{+ 0.30} & \textcolor{yellow!70!black}{+ 0.01} & \textcolor{yellow!70!black}{+ 0.04} \\
\cmidrule(lr){2-17}
\multirow{2}{*}{FactorizePhys}
  & UBFC-rPPG & - & - & - & - & - & \textcolor{yellow!70!black}{- 0.17} & \textcolor{yellow!70!black}{- 0.30} & \textcolor{yellow!70!black}{- 0.38} & \textcolor{yellow!70!black}{+ 0.01} & \textcolor{yellow!70!black}{- 0.11} & \textcolor{yellow!70!black}{- 0.05} & \textcolor{yellow!70!black}{- 0.01} & \textcolor{yellow!70!black}{- 0.06} & \textcolor{yellow!70!black}{0.00} & \textcolor{yellow!70!black}{- 0.10}\\
  & PURE      & \textcolor{yellow!70!black}{0.00} & \textcolor{yellow!70!black}{0.00} & \textcolor{yellow!70!black}{0.00} & \textcolor{yellow!70!black}{0.00} & \textcolor{yellow!70!black}{- 0.05} & - & - & - & - & - & \textcolor{green!70!black}{- 1.37} & \textcolor{green!70!black}{- 4.47} & \textcolor{green!70!black}{- 2.90} & \textcolor{green!70!black}{+ 0.13} & \textcolor{yellow!70!black}{- 0.07} \\
\bottomrule
\end{tabular}
}
\end{table}

\setlength{\intextsep}{0pt}
\begin{figure}[h]
    \centering
    \includegraphics[width=\linewidth]{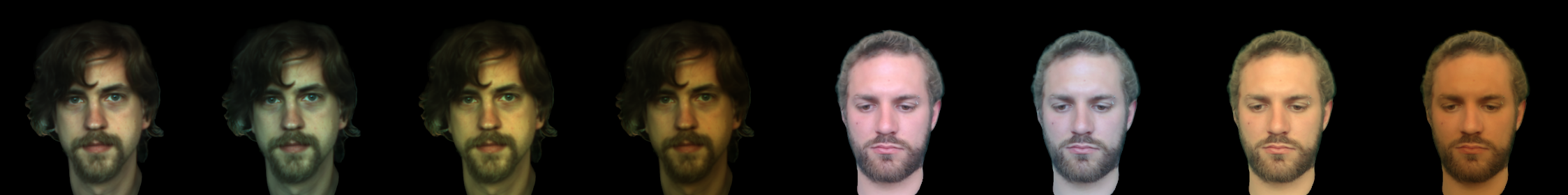}
    \vspace{-18pt}
    \caption{Relighting Ablation Visualization. From left to right: white, cool, warm, and real-world hospital lighting (\href{https://polyhaven.com/a/childrens_hospital}{Source Link}); two subjects from PURE (left) and UBFC-rPPG (right). Solid color environment maps are synthesized with matched luminance.}
    \label{fig:relit}
\end{figure}

\setlength{\intextsep}{6pt}
\begin{figure}[h]
    \centering
    \includegraphics[width=\linewidth]{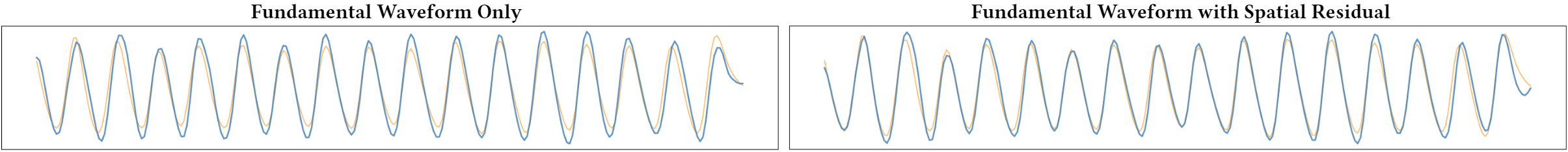}
    \vspace{-18pt}
    \caption{Ablation on Spatial Residual of Our Modulation Components. Both ours (blue) and GT (orange) are filtered and normalized. Full modulation shows a better alignment of the prescribed signals, with 0.05 MACC improved in this case.}
    \label{fig:placeholder}
\end{figure}


\setlength{\intextsep}{-5pt}
\begin{wrapfigure}{r}{0.2\columnwidth}
    \centering
    \includegraphics[width=0.18\columnwidth]{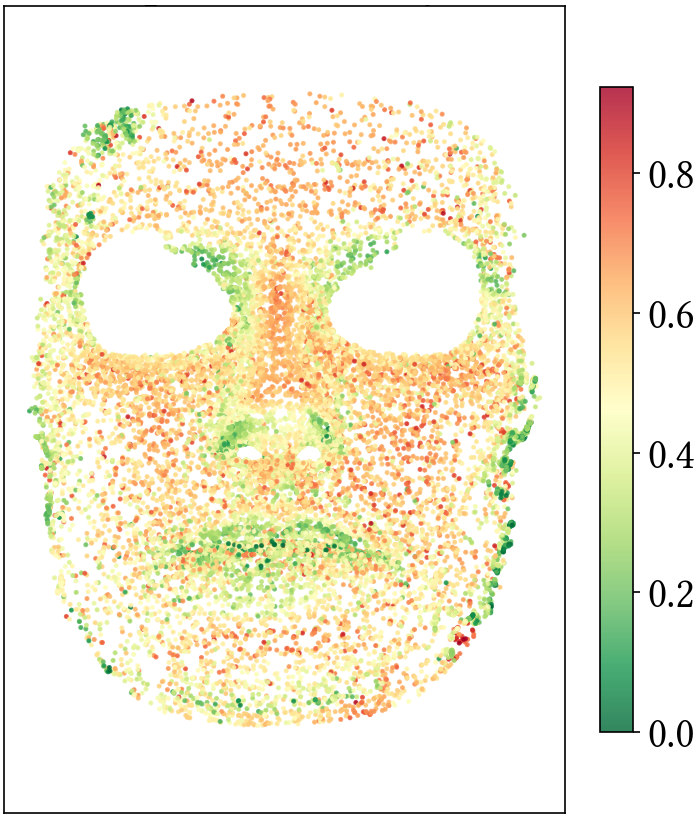}
    \caption{Mean spatial rPPG intensity across the facial skin region.}
    \label{fig:spatial}
\end{wrapfigure}

\vspace{-6pt}
\paragraph{Phase Parameterization}
We further experiment with regressing the direct phase coordinate $(x(t), y(t))$ on the unit circle per frame $t$ as an alternative. However, this approach empirically underperformed, as independent per-frame prediction lacks the temporal continuity that the cumulative formulation naturally enforces, making it harder to learn a coherent phase trajectory. While cumulative phase representations are theoretically susceptible to error accumulation across frames, our POS-based~\cite{POS2016} weak supervision acts as a periodicity anchor that continually corrects drift, mitigating the accumulation effect. Nevertheless, a drift-free phase representation across test frames remains a direction for future work.

\paragraph{Spatial Variation}
Our framework supports spatially varied rPPG signal intensity, as shown in Figure~\ref{fig:spatial}. However, it is not explicitly supervised - the pipeline supervises the mean green-channel albedo as the embedded signal. While the MLP implicitly learns non-uniform modulation patterns, the observed spatial structure remains strongly coupled to the underlying green channel albedo. Explicit spatial supervision could enable more principled control over per-region intensity, which we identify as a future direction.
\putbib[sample]
\end{bibunit}


\begin{thebibliography}{16}


\ifx \showCODEN    \undefined \def \showCODEN     #1{\unskip}     \fi
\ifx \showISBNx    \undefined \def \showISBNx     #1{\unskip}     \fi
\ifx \showISBNxiii \undefined \def \showISBNxiii  #1{\unskip}     \fi
\ifx \showISSN     \undefined \def \showISSN      #1{\unskip}     \fi
\ifx \showLCCN     \undefined \def \showLCCN      #1{\unskip}     \fi
\ifx \shownote     \undefined \def \shownote      #1{#1}          \fi
\ifx \showarticletitle \undefined \def \showarticletitle #1{#1}   \fi
\ifx \showURL      \undefined \def \showURL       {\relax}        \fi
\providecommand\bibfield[2]{#2}
\providecommand\bibinfo[2]{#2}
\providecommand\natexlab[1]{#1}
\providecommand\showeprint[2][]{arXiv:#2}

\bibitem[Bobbia et~al\mbox{.}(2019)]%
        {bobbia2017unsupervised}
\bibfield{author}{\bibinfo{person}{Serge Bobbia}, \bibinfo{person}{Richard Macwan}, \bibinfo{person}{Yannick Benezeth}, \bibinfo{person}{Alamin Mansouri}, {and} \bibinfo{person}{Julien Dubois}.} \bibinfo{year}{2019}\natexlab{}.
\newblock \showarticletitle{Unsupervised skin tissue segmentation for remote photoplethysmography}.
\newblock \bibinfo{journal}{\emph{Pattern Recognition Letters}} (\bibinfo{year}{2019}).
\newblock
\href{https://doi.org/10.1016/j.patrec.2017.10.017}{doi:\nolinkurl{10.1016/j.patrec.2017.10.017}}


\bibitem[Joshi et~al\mbox{.}(2024)]%
        {Fac2024}
\bibfield{author}{\bibinfo{person}{Jitesh Joshi}, \bibinfo{person}{Sos~S. Agaian}, {and} \bibinfo{person}{Youngjun Cho}.} \bibinfo{year}{2024}\natexlab{}.
\newblock \showarticletitle{FactorizePhys: Matrix Factorization for Multidimensional Attention in Remote Physiological Sensing}. In \bibinfo{booktitle}{\emph{Advances in Neural Information Processing Systems}}. \bibinfo{pages}{96607--96639}.
\newblock
\href{https://doi.org/10.52202/079017-3063}{doi:\nolinkurl{10.52202/079017-3063}}


\bibitem[Kerbl et~al\mbox{.}(2023)]%
        {kerbl3Dgaussians}
\bibfield{author}{\bibinfo{person}{Bernhard Kerbl}, \bibinfo{person}{Georgios Kopanas}, \bibinfo{person}{Thomas Leimk{\"u}hler}, {and} \bibinfo{person}{George Drettakis}.} \bibinfo{year}{2023}\natexlab{}.
\newblock \showarticletitle{3D Gaussian Splatting for Real-Time Radiance Field Rendering}.
\newblock \bibinfo{journal}{\emph{ACM Transactions on Graphics}} \bibinfo{volume}{42}, \bibinfo{number}{4} (\bibinfo{date}{July} \bibinfo{year}{2023}).
\newblock
\href{https://doi.org/10.48550/arXiv.2308.04079}{doi:\nolinkurl{10.48550/arXiv.2308.04079}}


\bibitem[Liu et~al\mbox{.}(2020)]%
        {TSCAN2020}
\bibfield{author}{\bibinfo{person}{Xin Liu}, \bibinfo{person}{Josh Fromm}, \bibinfo{person}{Shwetak Patel}, {and} \bibinfo{person}{Daniel McDuff}.} \bibinfo{year}{2020}\natexlab{}.
\newblock \showarticletitle{Multi-Task Temporal Shift Attention Networks for On-Device Contactless Vitals Measurement}. In \bibinfo{booktitle}{\emph{Advances in Neural Information Processing Systems}}. \bibinfo{pages}{19400--19411}.
\newblock


\bibitem[Liu et~al\mbox{.}(2023)]%
        {liu2022rppg}
\bibfield{author}{\bibinfo{person}{Xin Liu}, \bibinfo{person}{Girish Narayanswamy}, \bibinfo{person}{Akshay Paruchuri}, \bibinfo{person}{Xiaoyu Zhang}, \bibinfo{person}{Jiankai Tang}, \bibinfo{person}{Yuzhe Zhang}, \bibinfo{person}{Soumyadip Sengupta}, \bibinfo{person}{Shwetak Patel}, \bibinfo{person}{Yuntao Wang}, {and} \bibinfo{person}{Daniel McDuff}.} \bibinfo{year}{2023}\natexlab{}.
\newblock \bibinfo{title}{rPPG-Toolbox: Deep Remote PPG Toolbox}.
\newblock
\showeprint[arxiv]{2210.00716}


\bibitem[Lu et~al\mbox{.}(2025)]%
        {lu2025rhythmgaussian}
\bibfield{author}{\bibinfo{person}{Hao Lu}, \bibinfo{person}{Yuting Zhang}, \bibinfo{person}{Jiaqi Tang}, \bibinfo{person}{Bowen Fu}, \bibinfo{person}{Wenhang Ge}, \bibinfo{person}{Wei Wei}, \bibinfo{person}{Kaishun Wu}, {and} \bibinfo{person}{Yingcong Chen}.} \bibinfo{year}{2025}\natexlab{}.
\newblock \showarticletitle{RhythmGuassian: Repurposing Generalizable Gaussian Model For Remote Physiological Measurement}. In \bibinfo{booktitle}{\emph{Proceedings of the IEEE/CVF International Conference on Computer Vision (ICCV)}}.
\newblock


\bibitem[McDuff and Nowara(2021)]%
        {warmbodies}
\bibfield{author}{\bibinfo{person}{Daniel~J. McDuff} {and} \bibinfo{person}{Ewa~Magdalena Nowara}.} \bibinfo{year}{2021}\natexlab{}.
\newblock \showarticletitle{“Warm Bodies”: A Post-Processing Technique for Animating Dynamic Blood Flow on Photos and Avatars}.
\newblock \bibinfo{journal}{\emph{Proceedings of the 2021 CHI Conference on Human Factors in Computing Systems}} (\bibinfo{year}{2021}).
\newblock


\bibitem[Mildenhall et~al\mbox{.}(2020)]%
        {NeRF2020}
\bibfield{author}{\bibinfo{person}{Ben Mildenhall}, \bibinfo{person}{Pratul~P. Srinivasan}, \bibinfo{person}{Matthew Tancik}, \bibinfo{person}{Jonathan~T. Barron}, \bibinfo{person}{Ravi Ramamoorthi}, {and} \bibinfo{person}{Ren Ng}.} \bibinfo{year}{2020}\natexlab{}.
\newblock \showarticletitle{NeRF: Representing Scenes as Neural Radiance Fields for View Synthesis}. In \bibinfo{booktitle}{\emph{European Conference on Computer Vision (ECCV)}}.
\newblock


\bibitem[Paruchuri et~al\mbox{.}(2024)]%
        {MA2023}
\bibfield{author}{\bibinfo{person}{Akshay Paruchuri}, \bibinfo{person}{Xin Liu}, \bibinfo{person}{Yulu Pan}, \bibinfo{person}{Shwetak Patel}, \bibinfo{person}{Daniel McDuff}, {and} \bibinfo{person}{Soumyadip Sengupta}.} \bibinfo{year}{2024}\natexlab{}.
\newblock \showarticletitle{Motion Matters: Neural Motion Transfer for Better Camera Physiological Measurement}. In \bibinfo{booktitle}{\emph{Proceedings of the IEEE/CVF Winter Conference on Applications of Computer Vision}}. \bibinfo{pages}{5933--5942}.
\newblock
\href{https://doi.org/10.1109/WACV57701.2024.00583}{doi:\nolinkurl{10.1109/WACV57701.2024.00583}}


\bibitem[Stricker et~al\mbox{.}(2014)]%
        {PURE2014}
\bibfield{author}{\bibinfo{person}{Ronny Stricker}, \bibinfo{person}{Steffen M{\"u}ller}, {and} \bibinfo{person}{Horst-Michael Gross}.} \bibinfo{year}{2014}\natexlab{}.
\newblock \showarticletitle{Non-contact Video-based Pulse Rate Measurement on a Mobile Service Robot}. In \bibinfo{booktitle}{\emph{Proceedings of the 23rd IEEE International Symposium on Robot and Human Interactive Communication (RO-MAN 2014)}}. \bibinfo{pages}{1056--1062}.
\newblock
\href{https://doi.org/10.1109/ROMAN.2014.6926392}{doi:\nolinkurl{10.1109/ROMAN.2014.6926392}}


\bibitem[Tang et~al\mbox{.}(2023)]%
        {MMPD2023}
\bibfield{author}{\bibinfo{person}{Jiankai Tang}, \bibinfo{person}{Kequan Chen}, \bibinfo{person}{Yuntao Wang}, \bibinfo{person}{Yuanchun Shi}, \bibinfo{person}{Shwetak Patel}, \bibinfo{person}{Daniel McDuff}, {and} \bibinfo{person}{Xin Liu}.} \bibinfo{year}{2023}\natexlab{}.
\newblock \showarticletitle{MMPD: Multi-Domain Mobile Video Physiology Dataset}. In \bibinfo{booktitle}{\emph{2023 45th Annual International Conference of the IEEE Engineering in Medicine \& Biology Society (EMBC)}}. \bibinfo{pages}{1--5}.
\newblock
\href{https://doi.org/10.1109/EMBC40787.2023.10340857}{doi:\nolinkurl{10.1109/EMBC40787.2023.10340857}}


\bibitem[Tang et~al\mbox{.}(2020)]%
        {tang2020synthetic}
\bibfield{author}{\bibinfo{person}{Qunfeng Tang}, \bibinfo{person}{Zhencheng Chen}, \bibinfo{person}{Rabab Ward}, \bibinfo{person}{Carlo Menon}, {and} \bibinfo{person}{Mohamed Elgendi}.} \bibinfo{year}{2020}\natexlab{}.
\newblock \showarticletitle{Synthetic photoplethysmogram generation using two Gaussian functions}.
\newblock \bibinfo{journal}{\emph{Scientific Reports}} \bibinfo{volume}{10}, \bibinfo{number}{1} (\bibinfo{year}{2020}), \bibinfo{pages}{13883}.
\newblock
\href{https://doi.org/10.1038/s41598-020-69076-x}{doi:\nolinkurl{10.1038/s41598-020-69076-x}}


\bibitem[Wang et~al\mbox{.}(2016)]%
        {POS2016}
\bibfield{author}{\bibinfo{person}{Wenjin Wang}, \bibinfo{person}{Albertus~C. den Brinker}, \bibinfo{person}{Sander Stuijk}, {and} \bibinfo{person}{Gerard de Haan}.} \bibinfo{year}{2016}\natexlab{}.
\newblock \showarticletitle{Algorithmic Principles of Remote PPG}.
\newblock \bibinfo{journal}{\emph{IEEE Transactions on Biomedical Engineering}} \bibinfo{volume}{64}, \bibinfo{number}{7} (\bibinfo{year}{2016}), \bibinfo{pages}{1479--1491}.
\newblock
\href{https://doi.org/10.1109/TBME.2016.2609282}{doi:\nolinkurl{10.1109/TBME.2016.2609282}}


\bibitem[Wang et~al\mbox{.}(2022)]%
        {facevideo}
\bibfield{author}{\bibinfo{person}{Zhen Wang}, \bibinfo{person}{Yunhao Ba}, \bibinfo{person}{Pradyumna Chari}, \bibinfo{person}{Oyku~Deniz Bozkurt}, \bibinfo{person}{Gianna Brown}, \bibinfo{person}{Parth Patwa}, \bibinfo{person}{Niranjan Vaddi}, \bibinfo{person}{Laleh Jalilian}, {and} \bibinfo{person}{Achuta Kadambi}.} \bibinfo{year}{2022}\natexlab{}.
\newblock \showarticletitle{Synthetic Generation of Face Videos With Plethysmograph Physiology}. In \bibinfo{booktitle}{\emph{Proceedings of the IEEE/CVF Conference on Computer Vision and Pattern Recognition (CVPR)}}. \bibinfo{pages}{20587--20596}.
\newblock
\href{https://doi.org/10.1109/CVPR52688.2022.01993}{doi:\nolinkurl{10.1109/CVPR52688.2022.01993}}


\bibitem[Yu et~al\mbox{.}(2022)]%
        {Phys2022}
\bibfield{author}{\bibinfo{person}{Zitong Yu}, \bibinfo{person}{Yuming Shen}, \bibinfo{person}{Jingang Shi}, \bibinfo{person}{Hengshuang Zhao}, \bibinfo{person}{Philip Torr}, {and} \bibinfo{person}{Guoying Zhao}.} \bibinfo{year}{2022}\natexlab{}.
\newblock \showarticletitle{PhysFormer: Facial Video-based Physiological Measurement with Temporal Difference Transformer}. In \bibinfo{booktitle}{\emph{Proceedings of the Computer Vision and Pattern Recognition Conference (CVPR)}}.
\newblock
\href{https://doi.org/10.1109/CVPR52688.2022.00415}{doi:\nolinkurl{10.1109/CVPR52688.2022.00415}}


\bibitem[Zhang et~al\mbox{.}(2025)]%
        {zhang2024hravatar}
\bibfield{author}{\bibinfo{person}{Dongbin Zhang}, \bibinfo{person}{Yunfei Liu}, \bibinfo{person}{Lijian Lin}, \bibinfo{person}{Ye Zhu}, \bibinfo{person}{Kangjie Chen}, \bibinfo{person}{Minghan Qin}, \bibinfo{person}{Yu Li}, {and} \bibinfo{person}{Haoqian Wang}.} \bibinfo{year}{2025}\natexlab{}.
\newblock \showarticletitle{HRAvatar: High-Quality and Relightable Gaussian Head Avatar}. In \bibinfo{booktitle}{\emph{Proceedings of the Computer Vision and Pattern Recognition Conference (CVPR)}}. \bibinfo{pages}{26285--26296}.
\newblock
\href{https://doi.org/10.1109/CVPR52734.2025.02448}{doi:\nolinkurl{10.1109/CVPR52734.2025.02448}}


\end{thebibliography}
\end{document}